\documentclass{aims}
\usepackage{txfonts,cite}
\def\typeofarticle{Research article}
\def\currentvolume{31}
\def\currentissue{8}
\def\currentyear{2023}

\def\ppages{xxx--xxx}
\def\DOI{}
\def\Received{30 May 2023}
\def\Revised{21 June 2023}
\def\Accepted{25 June 2023}
\def\Published{}

\numberwithin{equation}{section}

\emergencystretch=\maxdimen
\makeatletter
\renewcommand{\@biblabel}[1]{#1\hfill \hspace{-0.2cm}}
\makeatother

\begin{document}

\title{Boundary distribution estimation for precise object detection}

\author{%
  Peng Zhi,
  Haoran Zhou,
  Hang Huang,
  Rui Zhao,
  Rui Zhou\corrauth~and
  Qingguo Zhou\corrauth
}

\shortauthors{the Author(s)}

\address{%
{School of Information Science and Engineering, Lanzhou University, Lanzhou, China}
  }

\corraddr{Email: zr@lzu.edu.cn, zhouqg@lzu.edu.cn; Tel: +869318912025; Fax: +869318912025.
}

\begin{abstract}
  In the field of state-of-the-art object detection, the task of object localization is typically accomplished through a dedicated subnet that emphasizes bounding box regression. This subnet traditionally predicts the object's position by regressing the box's center position and scaling factors. Despite the widespread adoption of this approach, we have observed that the localization results often suffer from defects, leading to unsatisfactory detector performance. In this paper, we address the shortcomings of previous methods through theoretical analysis and experimental verification and present an innovative solution for precise object detection. Instead of solely focusing on the object's center and size, our approach enhances the accuracy of bounding box localization by refining the box edges based on the estimated distribution at the object's boundary. Experimental results demonstrate the potential and generalizability of our proposed method.
\end{abstract}


\maketitle

\section{Introduction}

As a combination of classification and localization, object detection is intended to spot desired objects in the image and categorize them respectively. In recent years, the field of object classification has made significant advancements due to the progress in deep learning frameworks. Classifiers\cite{od-review, yolo-review, class1, class2} now demonstrate impressive performance on diverse and difficult benchmarks\cite{coco,imagenet, dataset1,pascal-voc}. However, the aspect of localization, which involves estimating the precise position and size of objects, still lags behind and poses limitations on the overall detection performance.

\begin{figure}[H]
  \begin{minipage}[b]{.48\linewidth}
    \centering
    \centerline{\includegraphics[width=6.0cm]{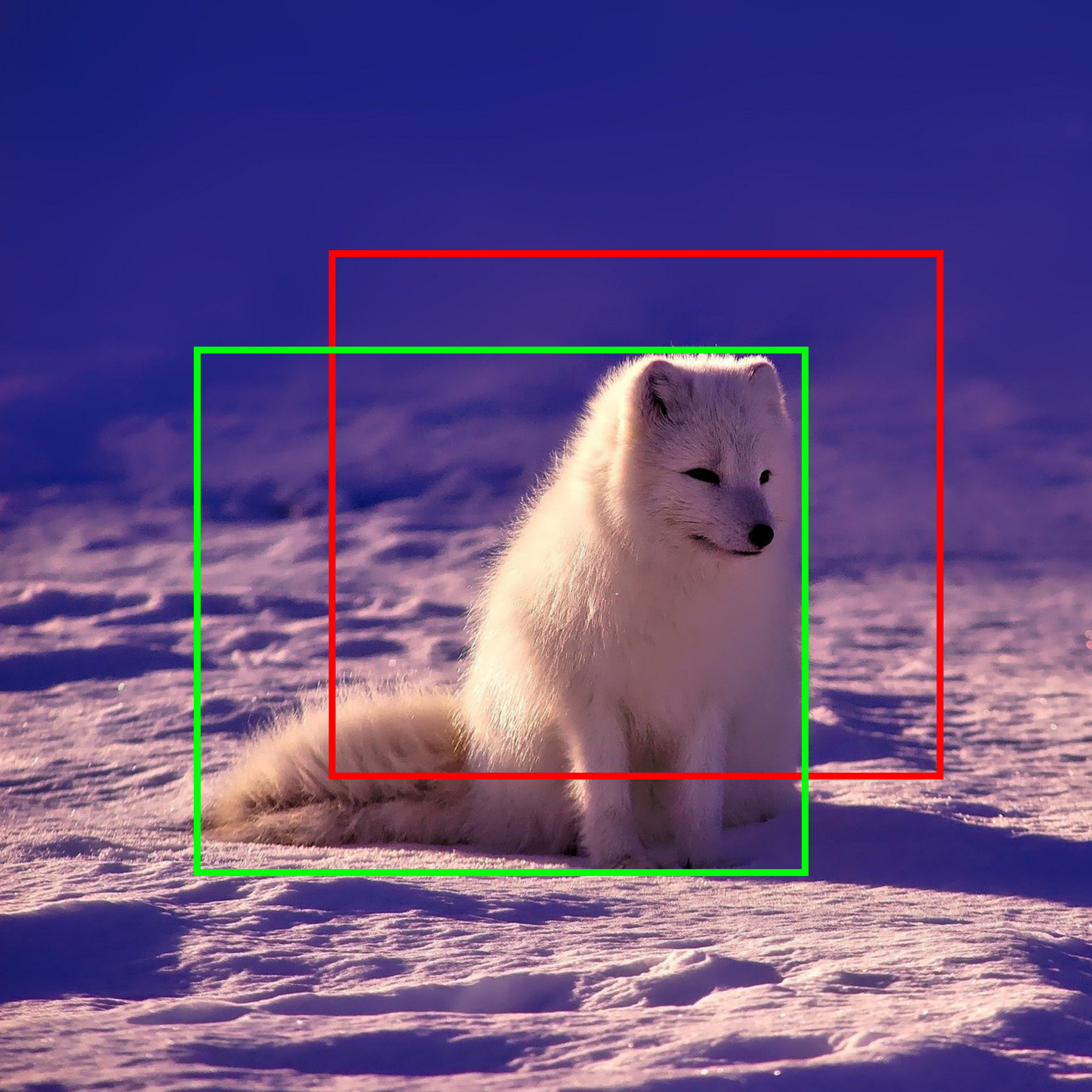}}
    \centerline{(a) Error in center position}\medskip
  \end{minipage}
  \hfill
  \begin{minipage}[b]{0.48\linewidth}
    \centering
    \centerline{\includegraphics[width=6.0cm]{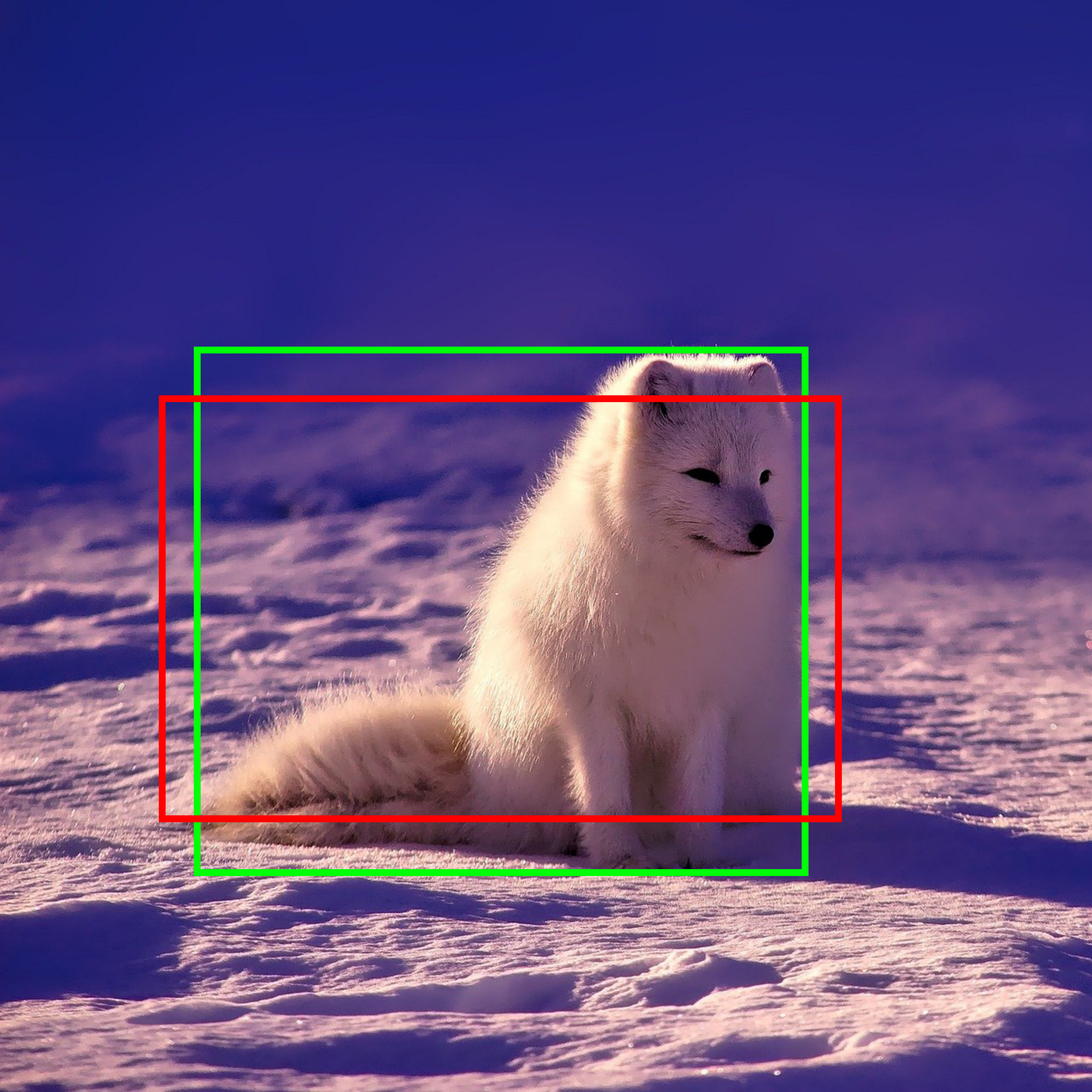}}
    \centerline{(b) Error in box size}\medskip
  \end{minipage}
  \caption{Illustration of the center-size box tuning method: the green box represents the ground truth, while the red box represents the predicted box with either center error or size~error.
  }
  \label{fig:fox_box}
\end{figure}

In state-of-the-art object detectors such as faster region-based convolutional neural network (R-CNN) \cite{faster-rcnn}, RetinaNet \cite{retinanet} and CenterNet \cite{centernet}, the task of object localization is carried out by a box subnet that focuses on bounding box regression. This subnet typically predicts bounding boxes by regressing the box's center position {$(x, y)$} and scaling factors {$(w, h)$}. While this design has shown effective results in conventional frameworks, it suffers from inherent flaws when it comes to precise localization. As illustrated in Figure~\ref{fig:fox_box}, each variable of {$(x, y, w, h)$} impacts multiple edges when adjusting the box's center position or size. Consequently, all edges of the box shift together, which compromises the accuracy of localization. Moreover, the center-focus representation method introduces a feature imbalance, wherein detectors disproportionately emphasize internal features that are less crucial for localization.

\begin{figure}[H]
    \centering
    \centerline{\includegraphics[width=15cm]{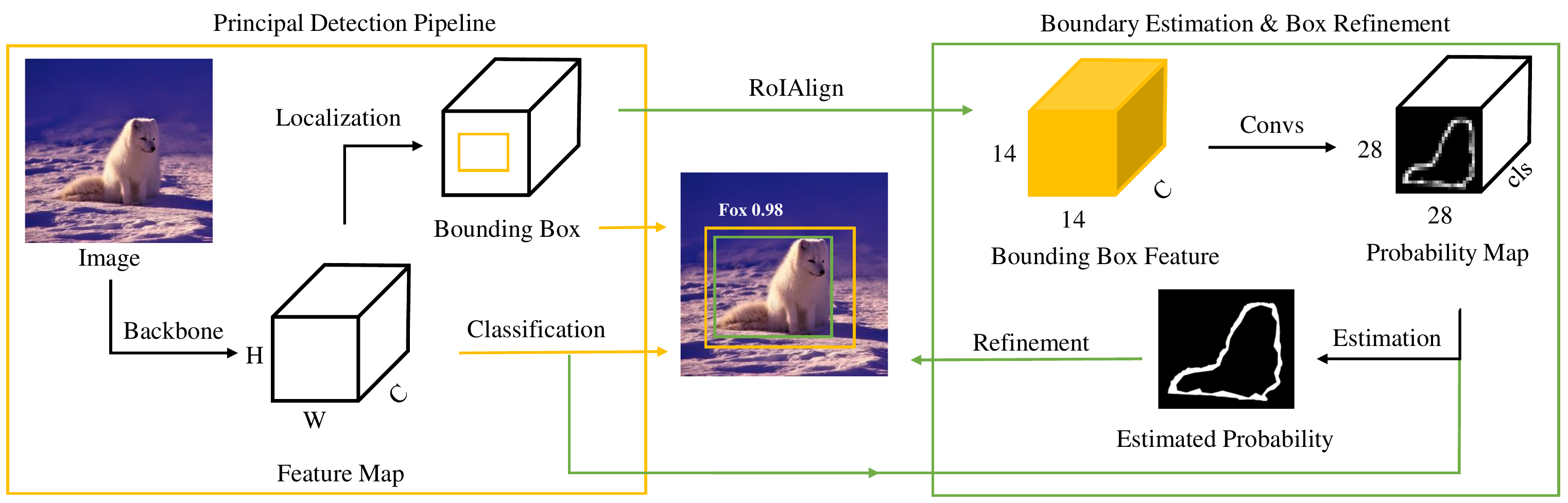}}
    \vspace{0.5cm}
  \caption{Illustration of principal detection pipelines with boundary estimation and box refinement.}
  \label{fig:pipeline}
\end{figure}

The aim of achieving precise localization is to minimize the disparity between predictions and the ground truth. When humans annotate ground truth boxes, they typically align each side of the box individually with the corresponding boundary, which is the most efficient approach. Taking inspiration from this manual annotation method, we introduce a novel solution for precise localization by refining previous localization results using estimations of the actual boundary distribution. Our proposed method follows a two-step process. As shown in Figure~\ref{fig:pipeline}, it first predicts the boundary within the proposed area of the existing pipeline. Next, the coarse boundary is determined based on the boundary probability map. For each side of the box, we estimate the corresponding fine boundary distribution by considering the context of the coarse boundary. This fine boundary distribution is then utilized to refine the final localization results.

We apply our proposed method to various frameworks and assess their performance on the common objects in context (COCO) test-dev dataset\cite{coco}. Our method demonstrates significant potential through straightforward functional estimation. Specifically, we achieved an improvement of 2.0\% in average precision (AP) without incurring additional computational costs or requiring additional annotation data, as compared to the competent Mask R-CNN \cite{mask-rcnn} baseline.

Our main contributions can be summarized as follows:
\begin{itemize}
	\item We propose a universal method for precise object detection that enhances the accuracy of bounding box localization by refining the box edges based on estimations of the object's boundary distribution. This novel approach improves the overall detection performance.
	\item In order to accurately generate each side of the bounding box, we evaluated various representation methods and conclude that the edge representation method is better suited for precise localization. This insight further enhances the effectiveness of our method.
	\item To leverage boundary features effectively, we introduce coarse-to-fine boundary distribution estimation modules. These modules demonstrate a significant improvement over previous cascade architectures, leading to more accurate and reliable object detection results.
\end{itemize}

\section{Related works}

\subsection{Localization in detection}

Due to the remarkable advancements in deep learning, convolutional neural network (CNN) based methods have emerged as the dominant solutions in detection applications. These deep learning detectors can be categorized into three main types based on their structures: two-stage detectors \cite{faster-rcnn,mask-rcnn,special-rcnn,ad-rcnn}, one-stage detectors \cite{yolo,ssd,redmon2018yolov3, pdnet} and anchor-free detectors \cite{centernet,fcos,3d-anchor}.

While different in their implementation, all of these methods adhere to the classification-localization paradigm. At its core, representing a bounding box requires four independent variables. In anchor-based detectors like Faster R-CNN \cite{faster-rcnn} and RetinaNet \cite{retinanet}, the bounding box is regressed from proposals or predefined anchors and represented as offsets of the center $(\delta x, \delta y)$ and relative scaling factors $(\delta w, \delta h)$. Similarly, anchor-free methods such as fully convolutional one-stage (FCOS) \cite{fcos} and CenterNet \cite{centernet} predict the object's center and its corresponding size. These center-size box tuning methods effectively locate objects but are still prone to inherent errors. To address this issue, CornerNet \cite{cornernet} introduces a novel approach by utilizing the object's top-left corner and bottom-right corner to form the bounding box. However, this method introduces additional computational costs due to corner pair grouping. In contrast, our method combines the strengths of these two approaches by refining the edges of previous localization results. By doing so, we can effectively enhance the accuracy of object detection without incurring the computational overhead associated with grouping corner pairs.

\subsection{Boundary in computer vision}

\begin{figure}[H]

  \begin{minipage}[b]{.23\linewidth}
    \centering
    \centerline{\includegraphics[width=2.8cm]{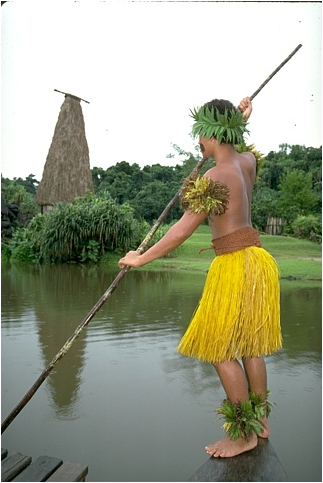}}
    \centerline{(a) original}\medskip
  \end{minipage}
  \hfill
  \begin{minipage}[b]{.23\linewidth}
    \centering
    \centerline{\includegraphics[width=2.8cm]{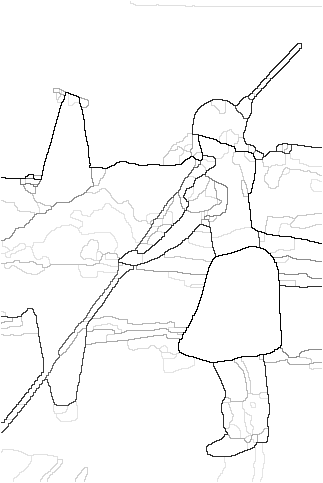}}
    \centerline{(b) edges}\medskip
  \end{minipage}
  \hfill
  \begin{minipage}[b]{.23\linewidth}
    \centering
    \centerline{\includegraphics[width=2.8cm]{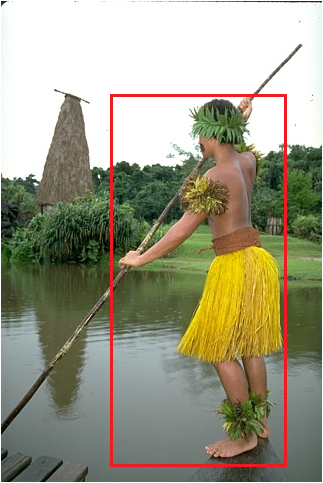}}
    \centerline{(c) bounding box}\medskip
  \end{minipage}
  \hfill
  \begin{minipage}[b]{.23\linewidth}
    \centering
    \centerline{\includegraphics[width=2.8cm]{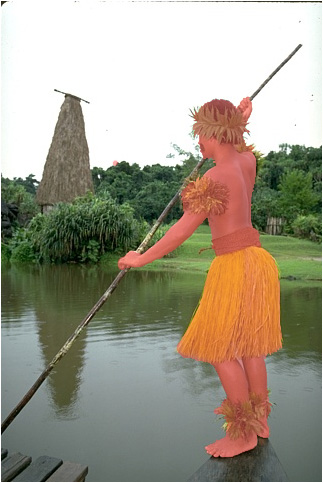}}
    \centerline{(d) mask}\medskip
  \end{minipage}
      \vspace{0.5cm}
  \caption{Illustration of some boundary-related applications in computer vision 
  (edge detection, object detection, and instance segmentation).}
  \label{fig:related}
\end{figure}

The semantic importance of edges and boundaries has long been recognized in various tasks at different levels. In Figure~\ref{fig:related}, we present two representative applications to highlight the similarities and differences between object detection and these tasks.

Edge detection, typically considered a low-level task, focuses on extracting visually prominent edges and object boundaries from natural images. It can be categorized into three main approaches: early filter methods \cite{canny1986a,marr1980theory,kittler1983on}, information theory methods \cite{martin2004learning, arbelaaez2011contour} and learning-based methods \cite{lim2013sketch,dollar2015fast}. Recent CNN-based methods \cite{xie2017holistically,bertasius2015deepedge,shen2015deepcontour} aim to automatically learn hierarchical features, but they still face challenges in accurately distinguishing object boundaries from detection results. Utilizing low-level cues, such as edges, to perform object-level tasks proves to be a challenging endeavor.

Instance segmentation aims to provide dense inference by generating pixel-wise masks for each instance of objects belonging to the same class. Some instance segmentation methods \cite{mask-rcnn,huang2019mask,chen2019hybrid} build upon existing detectors. They first utilize the detector to obtain object proposals and subsequently predict masks for these regions. However, these methods typically rely on pixel-level annotations, which are often unavailable in the context of object detection. While object detection shares some similarities with these tasks, such as the utilization of edges and boundaries, it also presents distinct challenges and requirements, such as precise localization, classification and the handling of multiple instances of objects. Overcoming these challenges often requires tailored approaches and adaptations specific to object detection scenarios.

     \vspace{0.5cm}

\subsection{Detection with boundary features}

Boundary context, including the surroundings and shapes of objects, contains valuable information that can benefit detection applications. Traditional detectors, such as Sobel \cite{kittler1983on}, Canny \cite{canny1986a}, and Marr-Hildreth \cite{marr1980theory}, typically rely on smoothing filters or image gradients to extract local features at a low level. Edge boxes \cite{edge-boxes} introduces a proposed algorithm based on edges in the input image, which works well for single objects but struggles to accurately separate multiple objects. These image-based detectors have limitations as they primarily consider the low-level visual context of objects and are challenging to apply to modern detectors. To address these limitations, recent deep learning-based methods have introduced novel techniques to incorporate context information. For example, the deformable R-FCN \cite{deformable-r-fcn} utilizes deformable convolution to include context information in object detection. The BAN \cite{ban} includes a boundary-aware network that leverages boundary context for improved performance. Side-aware boundary localization (SABL) \cite{sabl} aims to predict precise object boundaries using bucket-based approaches. Inspired by these deep learning-based methods, we referred to them and developed our estimation module to effectively leverage the boundary context of objects in our detection approach. By doing so, we enhance the overall accuracy and performance of object detection.

\section{Method}

In this research, our primary focus is to enhance the localization module in modern object detectors. Taking inspiration from the manual annotation method, we introduce an edge-focused box representation as a key component for achieving precise localization. Furthermore, we devise a novel boundary estimation method that effectively leverages boundary features. The overall pipeline of our proposed method is depicted in Figure~\ref{fig:pipeline}.

Our approach aligns with the fundamental detection pipeline, which considers object detection as a combination of localization and classification. Initially, features corresponding to the object are cropped and resized based on the previous localization results. These features are then utilized to predict the coarse boundary, which is subsequently refined to estimate the fine boundary. Each side of the bounding box is refined using the corresponding estimated fine boundary. Importantly, our method is not dependent on a specific detection pipeline and does not require any additional annotations. By employing this approach, we significantly improve the localization accuracy in object detection, without the need for extra annotations or modifications to the existing detection pipeline.

 \subsection{Bounding box representation}
The objective of precise localization is to minimize the discrepancy between predictions and the ground truth. In manual annotation, the most efficient approach is to align each side of the box with the object boundary individually. However, conventional detectors typically choose to predict the centers and sizes of bounding boxes instead of adjusting the edges directly.

As depicted in Figure~\ref{fig:fox_box}, the center-size tuning method inherently faces challenges when attempting to adjust individual edges. Let us consider a scenario in which each side of the box can be represented by its respective vertical and horizontal lines. The vertical and horizontal edges of the rectangular box $(E_l, E_r, E_t, E_b)$ can be represented by a set of scalars $(l ,r, t, b)$, or alternatively, by the box center and size $(x, y, w, h)$:
 \begin{equation}
 	E_l = f_l(l) = g_l(x, w/2)
 \end{equation}
 \begin{equation}
 	E_r = f_r(r) = g_r(x, w/2)
 \end{equation}
 \begin{equation}
 	E_t = f_t(t) = g_t(y, h/2)
 \end{equation}
 \begin{equation}
 	E_b = f_b(b) = g_b(y, h/2)
 \end{equation}
 
Each edge is influenced by both the center and size of the box. Consequently, adjusting a single edge becomes more difficult when employing the center-size method. This dilemma is prevalent in modern detectors, whether they predict the center and size directly or use the offset from the proposal relative to the ground truth box. A similar notion for precise localization is also reflected in the corner representation method \cite{cornernet}. The top-left corner and bottom-right corner can be viewed as the intersection points of the corresponding edges, denoted as $C_{tl}(l, t)$ and $C_{br}(r, b)$ respectively.
 
\subsection{Coarse boundary localization}
\label{sec:novel2}

\begin{figure}[H]
    \centering
    \centerline{\includegraphics[width=13cm]{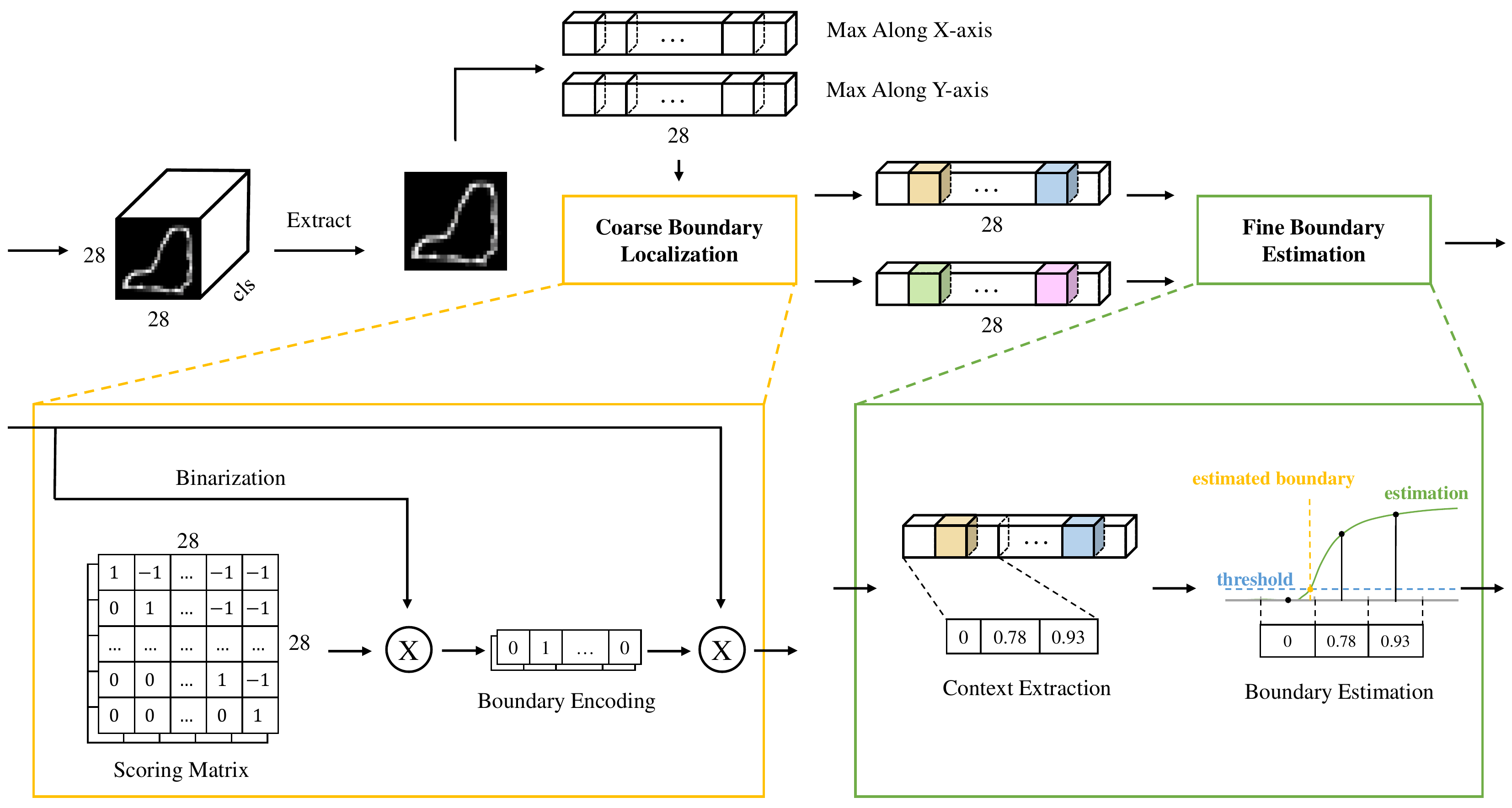}}
     \vspace{0.5cm}
  \caption{Illustration of part of the boundary estimation and box refinement module.
           }
  \label{fig:estimation}
\end{figure}

As depicted in Figures~\ref{fig:pipeline} and \ref{fig:estimation}, our method can be broken down into a two-step scheme: coarse boundary localization and fine boundary estimation. Initially, we utilize the region of interest align (RoIAlign) \cite{mask-rcnn} operation to extract and resize features within the proposal area. The resulting bounding box features, with dimensions of ${14 \times 14 \times C}$ (where ${C=256}$ in our experiments), are subjected to four ${3 \times 3}$ convolutional layers with a stride of 1, followed by a ${2 \times 2}$ deconvolutional layer with a stride of 2, and finally a ${1 \times 1}$ convolutional layer to produce the ${28 \times 28 \times cls}$ (where $cls=80$ in the COCO dataset \cite{coco}) boundary probability map. In conjunction with the classification result obtained from the previous detector, we extract a single-channel probability map for a specific class. Subsequently, we compute the maximum value along each row and column, resulting in two compressed vectors. These vectors serve as inputs to the coarse boundary localization module.
     \vspace{0.5cm}

Within the coarse boundary localization module, we first binarize the vector using a threshold (in our case, the threshold is set to 1 $\times$  $10^{-4} $) and multiply it by a scoring matrix and its transposed matrix. The scoring matrix, denoted as $M_s$ in Figure~\ref{fig:estimation}, undergoes matrix multiplication with the binarized vector $v_b$ according to the following equation:

 \begin{equation}
 	\vec{v_b} \times M_s = [\vec{v_b}\vec{m_{c1}}, ... ,\vec{v_b}\vec{m_{c28}}]
 \end{equation}
 And for each $\vec{v_b}\vec{m_{ci}}$, it follows that
 \begin{equation} 
 	\vec{v_b}\vec{m_{ci}} = \left\{
 	\begin{array}{ll}
 		0, & i < i_{fp}\\
 		1, & i = i_{fp}\\
 		1-k', & i_{fp} < i \le i_{lp}\\
 		-k, & i > i_{lp}
 	\end{array}
 	\right.
 \end{equation}
$i_{fp}$ and $i_{lp}$ represent the indices of the first and last positive elements in the vector, respectively. Additionally, $k$ and $k'$ are integers that satisfy $k,k' \in [0,28]$. These matrices are designed to identify the approximate positions of the left/top and right/bottom boundaries within the probability map. The resulting product is then activated using the rectified linear unit (ReLU) function.

 \subsection{Fine boundary estimation}

 As shown in Figure~\ref{fig:estimation}, the operation of the fine boundary estimation module begins by extracting the contextual information of the coarse boundary. The estimated fine boundary is then used to refine the bounding box. Assuming that the pixel values of the image are uniformly sampled from a continuous distribution, and that the variations near the object boundary can be described by an elementary function, we derive the precise boundary points based on this assumption and the pixel values near the object boundary. Our estimation method differs from the conventional interpolation approaches because it focuses on the distribution of the boundary rather than estimating values between known points. This allows us to select a relatively low threshold (in our experiments, we used a threshold of $0$) to minimize the outward expansion of the boundary during the interpolation process.

 For better understanding, let us assume that the one-dimensional vector obtained from the coarse-boundary localization process is denoted as \(V^{lr}\). By considering \(V^{lr}\) as a uniform sampling of a continuous distribution, with sampling points located at the pixel centers, we can transform the discrete \(V^{lr}\) into a continuous representation. Assuming that the distribution near the left boundary of the object can be approximated by an elementary function \(f(x)\), we utilize limit calculations and indices to demonstrate that the elementary function \(f(x)\) used to approximate the distribution near the boundary needs to pass through the coordinates \((0, 0)\) and \((1, 1)\) in the Cartesian coordinate system. The calculation process for the right boundary and the upper and lower boundaries of the target box follows a similar procedure, akin to the calculation based on \(V^{lr}\). Consequently, several key requirements must be fulfilled during the process. First, the boundary distribution function \(f_B: [0,1] \to [0,1]\) should be a monotonically increasing and differentiable function. Additionally, the gradient of the function should be smooth to alleviate numerical instability issues. For instance, the function \(f(x)=\sqrt{x}\) is unsuitable as a distribution function because its derivative \(f'(x)\) tends to infinity as \(x\) approaches \(0\), rendering it untrainable. In this study, we experimented with three elementary functions that meet the aforementioned requirements to fit the distribution of objects near the boundary, as detailed in Section \ref{sec:4-2}. Ultimately, a linear function was chosen as the fitting solution for this research.

 \section{Experiments}
 \subsection{Evaluation}

We present the evaluation results for our proposed estimation method, which were obtained by training and evaluating the networks on the COCO dataset \cite{coco}. To ensure fairness and consistency, we adhere to the universal splitting method for the training, validation and testing of our model on the COCO dataset. The detection results were evaluated using the standard COCO-style AP metric.

In our experiments, we chose Mask R-CNN \cite{mask-rcnn} as the baseline model. The parameters of the mask head in Mask R-CNN were used to initialize the network described in Section~\ref{sec:novel2}. This is because the mask head has an inherent advantage in predicting the boundary probability map. 

We observed that most elements inside the object are similar (close to 1), which renders them insignificant for estimation purposes. Consequently, we employed a functional estimation approach, denoted as $f(x)$, with the value $x$ at the coarse boundary serving as the variable.

\begin{figure}[H]
    \centering
    \centerline{\includegraphics[width=13cm]{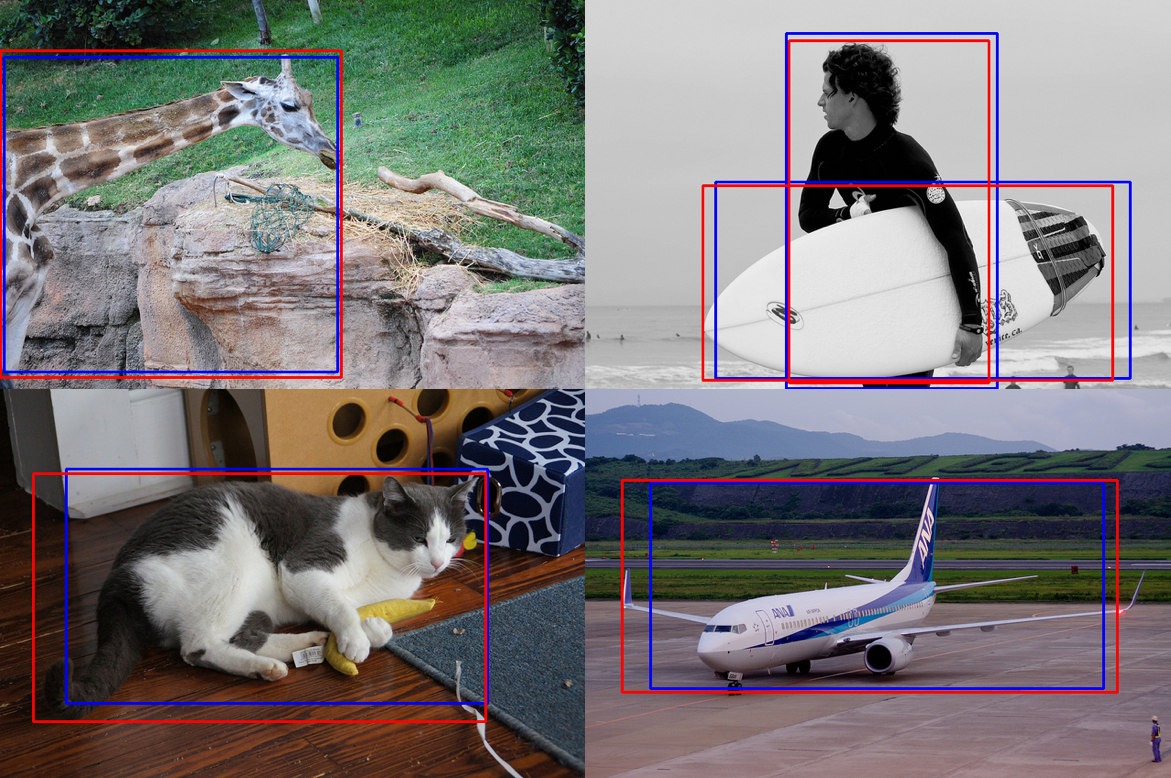}}
    \vspace{0.5cm}
  \caption{Detection results on the COCO test-dev dataset, with the baseline model shown in blue and our model shown in red.
           }
  \label{fig:coco-result}
\end{figure}

\begin{figure}[H]
    \centering
    \centerline{\includegraphics[width=15cm]{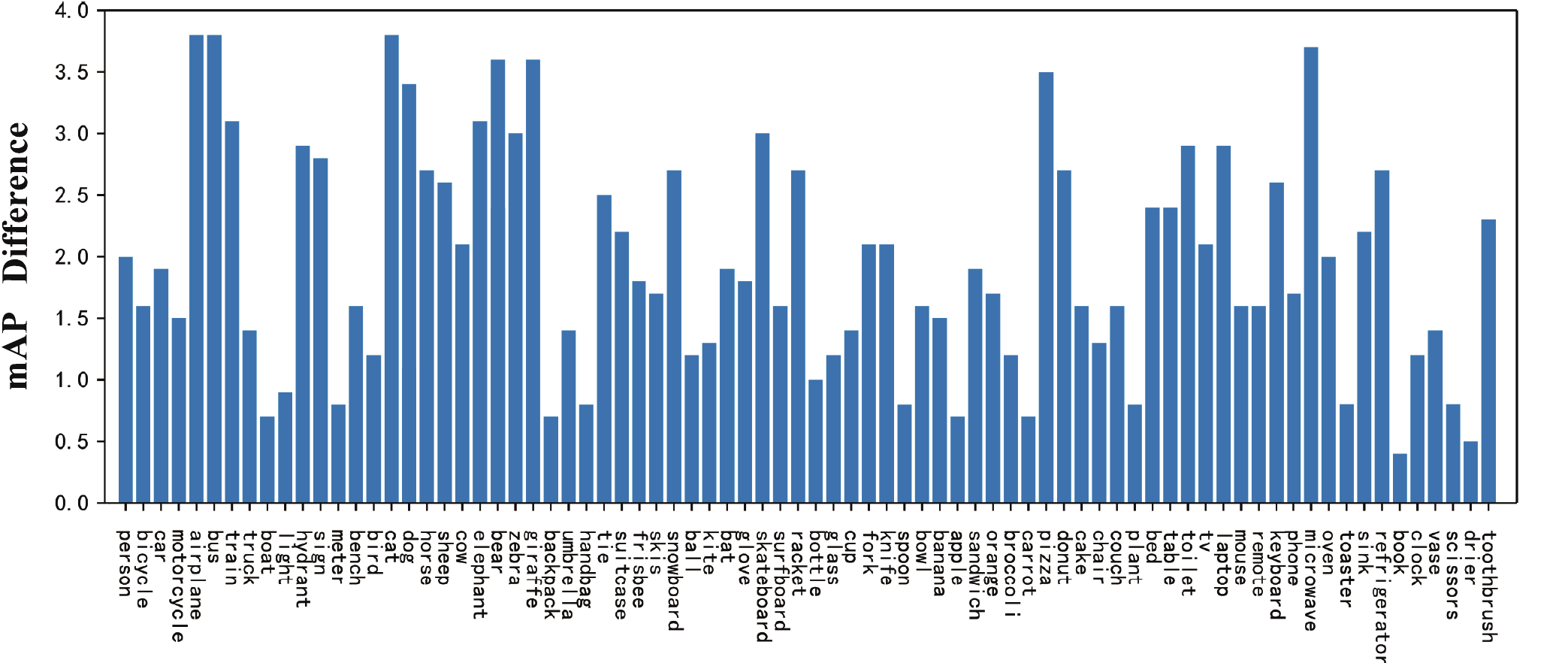}}
    \vspace{0.5cm}
  \caption{Illustration of the mean average precision (mAP) difference between our method and the baseline method on the COCO test-dev dataset for each category.
           }
  \label{fig:map-classes}
\end{figure}

The detection results are presented in Figure~\ref{fig:coco-result}, where the blue box represents the baseline method and the red box represents our method. As shown in Table~\ref{tab:eval} and Figure~\ref{fig:map-classes}, our method demonstrates improvements in overall AP, individual AP metrics and category-wise AP metrics as compared to the ResNet-50-FPN baseline with linear estimation $f(x)=x$. Notably, our method exhibits higher precision than the baseline even when both results are classified as positive examples. It is also noteworthy that our method outperforms the baseline when using a deeper ResNet-101-FPN backbone, despite the increased computational cost associated with this backbone.
     \vspace{0.5cm}

 \begin{table}[H]
  \begin{center}
    \caption{Comparison with other methods on COCO test-dev dataset.}
    \label{tab:eval}
  \begin{tabular}{llllllll}
  \hline
  Method                                   & Backbone       & AP   & $AP_{50}$ & $AP_{75}$ & $AP_S$ & $AP_M$ & $AP_L$ \\ \hline
  SSD513 \cite{ssd,fu2017dssd, mtdssd}      & ResNet-101     & 31.2 & 50.4      & 33.3      & 10.2   & 34.5   & 49.8   \\
  YOLOv3-608 \cite{redmon2018yolov3}       & Darknet-53     & 33.0 & 57.9      & 34.4      & 18.3   & 35.4   & 41.9   \\
  Faster R-CNN \cite{faster-rcnn}          & ResNet-101-FPN & 37.3 & 59.6      & 40.3      & 19.8   & 40.2   & 48.8   \\
  RetinaNet \cite{retinanet}               & ResNet-101-FPN & 39.1 & 59.1      & 42.3      & 21.8   & 42.7   & \bf{50.2}   \\
  Mask R-CNN \cite{mask-rcnn}              & ResNet-101-FPN & 38.2 & \bf{60.3}      & 41.7      & 20.1   & 41.1   & \bf{50.2}   \\
  RetinaMask \cite{fu2019retinamask}       & ResNet-50-FPN  & 39.4 & 58.6      & 42.3      & 21.9   & 42.0   & 51.0   \\ 
  Mask R-CNN (baseline)                     & ResNet-50-FPN  & 37.6 & 59.0      & 40.9      & 21.5   & 40.1   & 46.9   \\
  \bf{Ours}                                     & ResNet-50-FPN  & \bf{39.6} & 59.4      & \bf{42.5}      & \bf{22.2}   & \bf{42.2}   & 49.8   \\ \hline
  \end{tabular}
  \end{center}
  
\end{table}

     \vspace{0.5cm}

In addition, we conducted experiments using the pattern analysis, statical modeling and computational learning visual object classes (PASCAL VOC) dataset\cite{pascal-voc}. To expedite model convergence and reduce training costs, we fine-tuned the weights of the model pre-trained on the aforementioned COCO dataset. Subsequently, we evaluated the model's performance using the VOC 2012 validation set. The experimental results demonstrate the enhanced accuracy of our proposed method across various categories compared to Mask R-CNN\cite{mask-rcnn}. Specifically, our method achieves a notable improvement in the mAP metric, with an accuracy increase of 2.1 mAP $(48.0\rightarrow50.1)$. Additionally, we observed a significant precision gain of 2.8 on the AP for large objects (APL) metric $(56.3\rightarrow59.1)$. Notably, the \emph{airplane} category resulted in the largest improvement, with a precision gain of 3.5 mAP $(54.9\rightarrow58.4)$. Overall, our proposed method significantly outperforms Mask R-CNN on the PASCAL VOC dataset.

Apart from evaluating the accuracy, we thoroughly assessed the speed performance using frames per second (FPS) as a metric. To ensure reliable results, we conducted five measurements of FPS, with each measurement based on the processing time for 100 images of resolution $1024\times1024$. Mask R-CNN achieved an average FPS of 6.00, while our model achieved a slightly lower average FPS of 5.97, indicating a marginal 0.5\% reduction in FPS compared to Mask R-CNN. Therefore, the impact on speed can be considered negligible.

\subsection{Additional experiments and analysis}
\label{sec:4-2}
  {\bf Comparison between different estimation functions.}
In order to investigate the impact of different boundary distribution functions, we conducted comparative experiments with three cases (in Table~\ref{tab:func}): {\it linear}, {\it exponential} and {\it logarithmic}, represented by the functions $f(x)=x$, $f(x)=x^2$, and $f(x)=ln((e-1)x+1)$, respectively. From the experiment, we observed that the linear method achieved better overall performance, particularly on large objects. The convex logarithmic function performed better on smaller objects, exhibiting a higher $AP_{50}$ and ${AP_{75}}$. On the other hand, the concave exponential function did not demonstrate significant advantages.

\begin{table}[H]
  \begin{center}
  \caption{Comparison between different estimation functions on COCO val dataset.}
  \label{tab:func}
  \begin{tabular*}{\textwidth}{@{\extracolsep{\fill}}lllllll}
  \hline
  Method          & AP   & $AP_{50}$ & $AP_{75}$ & $AP_S$ & $AP_M$ & $AP_L$ \\ \hline
  Baseline        & 37.5 & 58.6      & 40.8      & 21.8   & 41.0   & 48.9   \\
  W/ lin          & \bf{39.5} & 59.0      & 42.4      & 22.4   & \bf{43.0}   & \bf{52.0}   \\
  W/ exp          & 39.3 & 58.9      & 41.9      & 22.4   & 42.9   & 51.7   \\
  W/ log          & 39.4 & \bf{59.2}      & \bf{42.6}      & \bf{22.8}   & \bf{43.0}   & 51.9   \\ \hline
  \end{tabular*}
\end{center}
\end{table}

  {\bf Influence of extra mask annotation.}
Although our method does not use annotation data during training, the initialization of the boundary prediction network with parameters from  Mask R-CNN \cite{mask-rcnn} introduces additional mask information. To assess the impact of mask annotation, we replaced the parameters in the boundary prediction network with randomly initialized parameters. As shown in Table~\ref{tab:param}, our method remains effective, with an overall AP improvement of 1.8\%. These results indicate that extra annotation, such as masks, can provide benefits but are not essential for the proposed method.

\begin{table}[H]
  \begin{center}   
  \caption{Ablation study on extra mask annotation.}
  \label{tab:param}
  \begin{tabular*}{\textwidth}{@{\extracolsep{\fill}}lllllll}
  \hline
  Method                             & AP   & $AP_{50}$ & $AP_{75}$ & $AP_S$ & $AP_M$ & $AP_L$ \\ \hline
  Baseline                         & 37.5 & 58.6      & 40.8      & 21.8   & 41.0   & 48.9   \\
  W/o mask & 38.5 & 58.3      & 41.1      & 21.6   & 42.3   & 50.5   \\
  W/  mask    & \bf{39.5} & \bf{59.0}      & \bf{42.4}      & \bf{22.4}   & \bf{43.0}   & \bf{52.0}   \\ \hline
  \end{tabular*}
  \end{center}
\end{table}

 {\bf Comparison with extra regression.}
 Cascading methods have demonstrated competence in precise localization. To highlight the superiority of our boundary estimation method, we compare it to baseline methods with an additional regression stage. As shown in Table~\ref{tab:reg}, our approach consistently outperforms the Mask R-CNN baselines, leading by up to 0.8\% in overall AP. These results validate the effectiveness of our method in achieving precise localization.
 
\begin{table}[H]
  \begin{center}
    \caption{Comparison between extra regression method and our edge refinement method.}
  \label{tab:reg}    
  \begin{tabular*}{\textwidth}{@{\extracolsep{\fill}}lllllll}
  \hline
  Method            & AP   & $AP_{50}$ & $AP_{75}$ & $AP_S$ & $AP_M$ & $AP_L$ \\ \hline
  Baseline       & 37.5 & 58.6      & 40.8      & 21.8   & 41.0   & 48.9   \\
  W/ reg      & 38.7 & 58.1      & 41.5      & 21.9   & 42.2   & 51.2   \\
  W/ ref & \bf{39.5} & \bf{59.0}      & \bf{42.4}      & \bf{22.4}   & \bf{43.0}   & \bf{52.0}   \\ \hline
  \end{tabular*}
  \end{center}
\end{table}

  \section{Conclusions}
  In this paper, we propose a novel approach to achieve precise object detection by estimating object boundary distributions. Our primary goal is to improve detection accuracy. Drawing inspiration from manual annotation techniques, we harness the advantages of the edge-focus box representation method. By predicting the coarse boundary and refining it through fine boundary estimation, our method significantly enhances the accuracy of bounding box edges based on previous localization results. Extensive experiments conducted on the COCO dataset demonstrate the potential of our boundary estimation method across various detection pipelines, eliminating the need for additional annotation. In future research, we aim to extend the application of our boundary detection procedure to diverse domains, including medical imaging. This extended application holds significant value, particularly in optical coherence tomography imaging, where our procedure enables the precise detection of boundaries in retina tissue layers with high accuracy. This advancement has the potential to facilitate the accurate diagnosis of ocular or neuropathological disorders \cite{RKM1,RKM2}. Furthermore, we plan to explore the utilization of a lighter ResNet18 backbone or a more complex ResNet101 backbone to address the specific requirements of speed and accuracy in different scenarios.

\section*{Acknowledgments}

This work was partially supported by the National Key R\&D Program of China under Grant No. 2020YFC0832500, Gansu Province Science and Technology Major Project - Industrial Project under Grant No. 22ZD6GA048, Gansu Province Key Research and Development Plan - Industrial Project under Grant No. 22YF7GA004, Fundamental Research Funds for the Central Universities under Grant Nos. lzujbky-2022-kb12, lzujbky-2021-sp43, lzujbky-2020-sp02, lzujbky-2019-kb51 and lzujbky-2018-k12, National Natural Science Foundation of China under Grant No. 61402210, Science and Technology Plan of Qinghai Province under Grant No.2020-GX-164 and Supercomputing Center of Lanzhou University. We also gratefully acknowledge the support of NVIDIA Corporation for the donation of the Jetson TX1 used for this research. This work was also partially supported by the Gansu Provincial Science and Technology Major Special Innovation Consortium Project (21ZD3GA002), from the Gansu Province Green and Smart Highway Transportation Innovation Consortium as part of the Gansu Province Green and Smart Highway Key Technology Research and Demonstration. We would like to express our gratitude to our co-authors, Mr. Hang Huang and Mr. Haoran Zhou, for their dedicated efforts during their postgraduate studies\cite{huanghang}. We acknowledge their hard work and appreciate their significant contributions to the research.


\begin{thebibliography}{999}

\bibitem{od-review}
R. Kaur, S. Singh,
A comprehensive review of object detection with deep learning,
\emph{Digital Signal Process.}, \textbf{132} (2023), 103812.
\doilink{https://doi.org/10.1016/j.dsp.2022.103812}

\bibitem[2]{yolo-review}
P. Jiang, D. Ergu, F. Liu, Y. Cai, B. Ma,
A Review of Yolo algorithm developments,
\emph{Proc. Comput. Sci.}, \textbf{199} (2022), 1066--1073.
    \doilink{https://doi.org/10.1016/j.procs.2022.01.135}

\bibitem[3]{class1}
W. Liu, G. Wu, F. Ren, X. Kang,
DFF-ResNet: An insect pest recognition model based on residual networks,
\emph{Big Data Min. Anal.}, \textbf{3} (2020), 300--310.
\doilink{https://doi.org/10.26599/BDMA.2020.9020021}

\bibitem[4]{class2}
A. Mughees, L. Tao,
Multiple deep-belief-network-based spectral-spatial classification of hyperspectral images,
\emph{Tsinghua Sci. Technol.}, \textbf{24} (2019), 183--194.
\doilink{https://doi.org/10.26599/TST.2018.9010043}

\bibitem[5]{coco}
T. Y. Lin, M. Maire, S. Belongie, J. Hays, P. Perona, D. Ramanan, et al., 
Microsoft COCO: Common objects in context,
in \emph{European Conference on Computer Vision}, (2014), 740--755.
    \doilink{https://doi.org/10.1007/978-3-319-10602-1\_48}

\bibitem[6]{imagenet}
O. Russakovsky, J. Deng, H. Su, J. Krause, S. Satheesh, S. Ma, et al., 
ImageNet large scale visual recognition challenge, 
\emph{Int. J. Comput. Vis.}, \textbf{115} (2015), 211--252.
\doilink{https://doi.org/10.1007/s11263-015-0816-y}

\bibitem[7]{dataset1}
Y. Fan, D. Ni, H. Ma,
HyperDB: a hyperspectral land class database designed for an image processing system,
\emph{Tsinghua Sci. Technol.}, \textbf{22} (2017), 112--118.
\doilink{https://doi.org/10.1109/TST.2017.7830901}

\bibitem[8]{pascal-voc}
 M. Everingham, S. M. A. Eslami, L. Van Gool, C. K. I. Williams, J. Winn, A. Zisserman ,
The PASCAL visual object classes challenge: A retrospective,
\emph{Int. J. Comput. Vis.}, \textbf{111} (2015), 98--136.
\doilink{https://doi.org/10.1007/s11263-014-0733-5}

\bibitem[9]{faster-rcnn}
S. Ren, K. He, R. Girshick, J. Sun,
 Faster R-CNN: Towards real-time object detection with region proposal networks,
\emph{IEEE Trans. Pattern Anal. Mach. Intell.}, \textbf{39} (2017), 1137-1149.
\doilink{https://doi.org/10.1109/TPAMI.2016.2577031}

\bibitem[10]{retinanet}
T. Y. Lin, P. Goyal, R. Girshick, K. He, P. Dollár,
Focal loss for dense object detection,
in \emph{2017 IEEE International Conference on Computer Vision (ICCV)}, (2017), 2999--3007.
    \doilink{https://doi.org/10.1109/ICCV.2017.324}

\bibitem[11]{centernet}
 X. Zhou, D. Wang, P. Krähenbühl,
Objects as points,
preprint, arXiv:1904.07850.

\bibitem[12]{mask-rcnn}
K. He, G. Gkioxari, P. Dollár, R. Girshick,
Mask R-CNN, 
\emph{IEEE Trans. Pattern Anal. Mach. Intell.}, \textbf{42} (2020), 386--397.
\doilink{https://doi.org/10.1109/TPAMI.2018.2844175}

\bibitem[13]{special-rcnn}
M. Chen, F. Bai, Z. Gerile,
Special object detection based on Mask RCNN,
in \emph{2021 17th International Conference on Computational Intelligence and Security (CIS)}, (2021), 128--132.
\doilink{https://doi.org/10.1109/CIS54983.2021.00035}

\bibitem[14]{ad-rcnn}
Z. Ou, Z. Wang, F. Xiao, B. Xiong, H. Zhang, M. Song, et al.,
AD-RCNN: Adaptive dynamic neural network for small object detection,
\emph{IEEE Int. Things J.}, \textbf{10} (2023), 4226--4238.
\doilink{https://doi.org/10.1109/JIOT.2022.3215469}

\bibitem[15]{pdnet}
 L. Yang, Y. Xu, S. Wang, C. Yang, Z. Zhang, B. Li, et al.,
PDNet: Toward better one-stage object detection with prediction decoupling,
\emph{IEEE Trans. Image Process.}, \textbf{31} (2022), 5121--5133.
\doilink{https://doi.org/10.1109/TIP.2022.3193223}

\bibitem[16]{yolo}
J. Redmon, S. Divvala, R. Girshick, A. Farhadi,
You only look once: Unified, real-time object detection,
 in \emph{2016 IEEE Conference on Computer Vision and Pattern Recognition (CVPR)}, (2016), 770--778.
    \doilink{https://doi.ieeecomputersociety.org/10.1109/CVPR.2016.91}

\bibitem[17]{ssd}
W. Liu, D. Anguelov, D. Erhan, C. Szegedy, S. Reed, C. Y. Fu, et al.,
SSD: Single shot multiBox detector,
in \emph{European Conference on Computer Vision}, (2016), 21--37.
    \doilink{https://doi.org/10.1007/978-3-319-46448-0\_2}

\bibitem[18]{redmon2018yolov3}
J. Redmon, A. Farhadi,
YOLOv3: An incremental improvement,
preprint, arXiv:1804.02767.

\bibitem[19]{3d-anchor}
G. Wang, J. Wu, B. Tian, S. Teng, L. Chen, D. Cao, et al.,
CenterNet3D: An anchor free object detector for point cloud,
\emph{IEEE Trans. Intell. Transp. Syst.}, \textbf{23} (2022), 12953--12965.
\doilink{https://doi.org/10.1109/TITS.2021.3118698}

\bibitem[20]{fcos}
Z. Tian, C. Shen, H. Chen, T. He,
FCOS: Fully convolutional one-stage object detection,
in \emph{2019 IEEE/CVF International Conference on Computer Vision (ICCV)},  (2019), 9626--9635.
 \doilink{https://doi.ieeecomputersociety.org/10.1109/ICCV.2019.00972}

\bibitem[21]{cornernet}
H. Law, J. Deng,
CornerNet: Detecting objects as paired keypoints,
\emph{Int. J. Comput. Vis.}, \textbf{128} (2020), 642--656.
\doilink{https://doi.org/10.1007/s11263-019-01204-1}

\bibitem[22]{canny1986a}
J. Canny, 
A computational approach to edge detection,
\emph{IEEE Trans. Pattern Anal. Mach. Intell.}, \textbf{8} (1986), 679--698.
\doilink{https://doi.org/10.1109/TPAMI.1986.4767851}

\bibitem[23]{marr1980theory}
D. Marr, E. Hildreth,
Theory of edge detection,
\emph{Proc. R. Soc. Lond. B}, \textbf{207} (1980), 187--217.
\doilink{https://doi.org/10.1098/rspb.1980.0020}

\bibitem[24]{kittler1983on}
J. Kittler,
On the accuracy of the Sobel edge detector, \emph{Image Vis. Comput.}, \textbf{1} (1983), 37--42.
\doilink{https://doi.org/10.1016/0262-8856(83)90006-9}

\bibitem[25]{martin2004learning}
D. R. Martin, C. C. Fowlkes, J. Malik,
Learning to detect natural image boundaries using local brightness, color, and texture cues,
\emph{IEEE Trans. Pattern Anal. Mach. Intell.}, \textbf{26} (2004), 530--549.
\doilink{https://doi.org/10.1109/TPAMI.2004.1273918}

\bibitem[26]{arbelaaez2011contour}
P.  Arbeláez, M. Maire, C. Fowlkes, J. Malik,
Contour detection and hierarchical image segmentation,
\emph{IEEE Trans. Pattern Anal. Mach. Intell.}, \textbf{33} (2011), 898--916.
\doilink{https://doi.org/10.1109/TPAMI.2010.161}

\bibitem[27]{lim2013sketch}
J. J. Lim, C. L. Zitnick, P. Dollár, 
Sketch tokens: A learned mid-level representation for contour and object detection,
in \emph{2013 IEEE Conference on Computer Vision and Pattern Recognitionn},  (2013), 3158--3165.
\doilink{https://doi.org/10.1109/CVPR.2013.406}

\bibitem[28]{dollar2015fast}
P. Dollár, C. L. Zitnick,
Structured forests for fast edge detection,
in \emph{ 2013 IEEE International Conference on Computer Vision}, (2013), 1841--1848.
 \doilink{https://doi.org/10.1109/ICCV.2013.231}

\bibitem[29]{xie2017holistically}
S. Xie, Z. Tu,
 Holistically-nested edge detection,
  in \emph{2015 IEEE International Conference on Computer Vision (ICCV)}, (2015), 1395--1403.
    \doilink{https://doi.org/10.1109/ICCV.2015.164}

\bibitem[30]{bertasius2015deepedge}
G. Bertasius, J. Shi, L. Torresani,
DeepEdge: A multi-scale bifurcated deep network for top-down contour detection,
in \emph{2015 IEEE Conference on Computer Vision and Pattern Recognition (CVPR)}, (2015), 4380--4389.
    \doilink{https://doi.org/10.1109/CVPR.2015.7299067}

\bibitem[31]{shen2015deepcontour}
W. Shen, X. Wang, Y. Wang, X. Bai, Z. Zhang,
DeepContour: A deep convolutional feature learned by positive-sharing loss for contour detection,
in \emph{2015 IEEE Conference on Computer Vision and Pattern Recognition (CVPR)}, (2015), 3982--3991.
    \doilink{https://doi.org/10.1109/CVPR.2015.7299024}

\bibitem[32]{huang2019mask}
Z. Huang, L. Huang, Y. Gong, C. Huang, X. Wang,
Mask scoring R-CNN,
in \emph{2019 IEEE/CVF Conference on Computer Vision and Pattern Recognition (CVPR)}, (2019), 6402--6411.
    \doilink{https://doi.org/10.1109/CVPR.2019.00657}

\bibitem[33]{chen2019hybrid}
K. Chen, J. Pang, J. Wang, Y. Xiong, X. Li, S. Sun, et al.,
  Hybrid task cascade for instance segmentation,
in \emph{ 2019 IEEE/CVF Conference on Computer Vision and Pattern Recognition (CVPR)}, (2019), 4974--4983.
    \doilink{https://doi.org/10.1109/CVPR.2019.00511}

\bibitem[34]{edge-boxes}
C. L. Zitnick, P. Dollár,
Edge boxes: Locating object proposals from edges,
in \emph{European Conference on Computer Vision}, (2014), 391--405.
    \doilink{https://doi.org/10.1007/978-3-319-10602-1\_26}

\bibitem[35]{deformable-r-fcn}
   J. Dai, H. Qi, Y. Xiong, Y. Li, G. Zhang, H. Hu, et al.,
  Deformable convolutional networks,
  in \emph{2017 IEEE International Conference on Computer Vision (ICCV)}, (2017), 764--773.
    \doilink{https://doi.org/10.1109/ICCV.2017.89}


\bibitem[36]{ban}
 Y. Kim, T. Kim, B. N. Kang, J. Kim, D. Kim,
  BAN: Focusing on boundary context for object detection,
  in \emph{Asian Conference on Computer Vision}, 2018, 555--570.
    \doilink{https://doi.org/10.1007/978-3-030-20876-9\_35}

\bibitem[37]{sabl}
J. Wang, W. Zhang, Y. Cao, K. Chen, J. Pang, T. Gong, et al.,
Side-aware boundary localization for more precise object detection,
in \emph{European Conference on Computer Vision}, (2020), 403--419.
\doilink{https://doi.org/10.1007/978-3-030-58548-8\_24}

\bibitem[38]{fu2017dssd}
 C. Y. {Fu}, W. {Liu}, A. {Ranga}, A. {Tyagi}, A. C. {Berg},
DSSD: Deconvolutional single shot detector,
preprint, 	arXiv:1701.06659.

\bibitem[39]{mtdssd}
R. Araki, T. Onishi, T. Hirakawa, T. Yamashita, H. Fujiyoshi,
    MT-DSSD: Deconvolutional single shot detector using multi task learning for object detection, segmentation, and grasping detection,
in \emph{ 2020 IEEE International Conference on Robotics and Automation (ICRA)}, (2020),  10487--10493.
    \doilink{https://doi.org/10.1109/ICRA40945.2020.9197251}

\bibitem[40]{fu2019retinamask}
 C. Y. Fu, M. Shvets, A. C. Berg, 
  RetinaMask: Learning to predict masks improves state-of-the-art single-shot detection for free,
preprint, arXiv:1901.03353.

\bibitem[41]{RKM1}
R. K. Meleppat, M. V. Matham, L. K. Seah,
 Optical frequency domain imaging with a rapidly swept laser in the 1300nm bio-imaging window,
  in \emph{ International Conference on Optical and Photonic Engineering},  (2015), 721--729. 
    \doilink{https://doi.org/10.1117/12.2190530}

\bibitem[42]{RKM2}
  R. K. Meleppat, C. R. Fortenbach, Y. Jian, E. S. Martinez, K. Wagner, B. S. Modjtahedi, et al.,
In Vivo Imaging of Retinal and Choroidal Morphology and Vascular Plexuses of Vertebrates Using Swept-Source Optical Coherence Tomography,
 \emph{Transl. Vis. Sci. Technol.}, \textbf{11} (2022), 11. 
    \doilink{https://doi.org/10.1167/tvst.11.8.11}

\bibitem[43]{huanghang}
H. Huang, 
\emph{Research on Object Detection Based on Improved MASK R-CNN},
Master's degree, Lanzhou University in Lanzhou, 2021.
    \doilink{https://doi.org/10.27204/d.cnki.glzhu.2021.001818}

\end{thebibliography}
\end{document}